\pdfoutput=1

\documentclass[11pt]{article}

\usepackage[]{emnlp2021}
\usepackage{amsmath}
\usepackage{times}
\usepackage{latexsym}
\usepackage{paralist}
\usepackage{multirow}
\usepackage[T1]{fontenc}

\usepackage[utf8]{inputenc}

\usepackage{microtype}

\usepackage{graphicx}
\usepackage{amsmath}

\title{Understanding ME? \\
Multimodal Evaluation for Fine-grained Visual Commonsense}

\author{Zhecan Wang\textsuperscript{1}, Haoxuan You\textsuperscript{1}, Yicheng He\textsuperscript{1}, Wenhao Li\textsuperscript{1}, Kai-Wei Chang\textsuperscript{2}, Shih-Fu Chang\textsuperscript{1}
\\
\textsuperscript{1} Columbia University, New York, \textsuperscript{2} University of California, Los Angeles \\
  \texttt{\{zw2627, hy2612, yh3330, wl2750, sc250\}@columbia.edu, kwchang@cs.ucla.edu} 
  }

\begin{document}
\maketitle
\begin{abstract}


 Visual commonsense understanding requires Vision Language (VL) models to not only understand image and text but also cross-reference in-between to fully integrate and achieve comprehension of the visual scene described. Recently, various approaches have been developed and have achieved high performance on visual commonsense benchmarks. However, it is unclear whether the models really understand the visual scene and underlying commonsense knowledge due to limited evaluation data resources. To provide an in-depth analysis, we present a Multimodal Evaluation (ME) pipeline to automatically generate question-answer pairs to test models' understanding of the visual scene, text, and related knowledge. We then take a step further to show that training with the ME data boosts model's performance in standard VCR evaluation. Lastly, our in-depth analysis and comparison  reveal interesting findings: (1) semantically low-level information can assist learning of high-level information but not the opposite; (2) visual information is generally under utilization compared with text.

\end{abstract}

\begin{figure}[ht!]
\begin{center}
\scalebox{0.75}{
  \includegraphics[width=\linewidth]{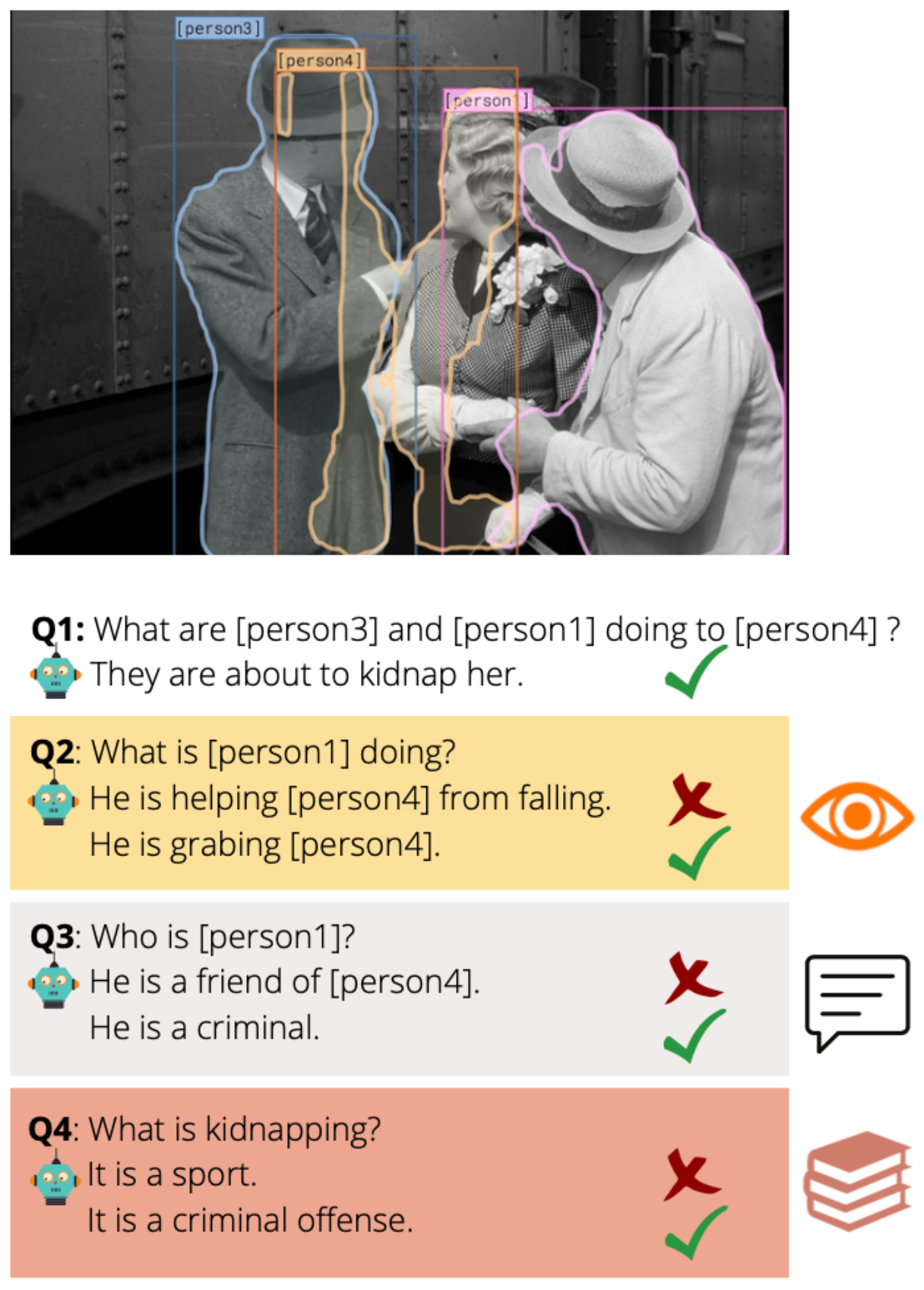}
}
\end{center}
  \caption{An example from VL benchmark, Visual Commonsense Reasoning (VCR) \cite{vcr}. VL models can answer the highly semantic VCR question correctly but fail terribly in answering related visual question (Q2), textual question (Q3), and background knowledge question (Q4).}
\label{fig:motivation}
\end{figure}

\section{Introduction}


Vision Language (VL) understanding is challenging because it requires VL models to identify and integrate information from both modalities to fully understand visual scenes. Numerous VL benchmarks have been created such as CLEVER \cite{johnson2017clevr}, GQA \cite{hudson2019gqa}, VQA \cite{li2018vqa}, VCR \cite{vcr} and SNLI-VE \cite{xie2018visual}. These benchmarks typically form VL evaluation in question-answering format with images and test models' understanding of both VL modalities. Despite the high accuracy achieved by existing large pretrained VL models, recent works have pointed out that VL models tend to exploit data biases such as shallow mappings with language priors and unbalanced utilization of information between modalities \cite{cao2020behind, jimenez2022carets, selvaraju2020squinting}.




 As in Fig. \ref{fig:motivation}, notwithstanding the model's success in answering Q1, the same model fails to answer the related visual, textual, and background questions. This example demonstrates that the VL model does not fully understand the visual scene, which leads to prediction inconsistency (when one model makes conflicting(inconsistent) predictions in two related questions). 
In our analysis, prediction inconsistency is surprisingly common among models in different modalities. 
Former works have also pointed out that most VQA systems achieve only middling self-consistency ($60-80 \%$) \cite{jimenez2022carets}. Therefore, 
we pose doubts to existing VL models' ability to thoroughly comprehend visual commonsense 
despite their high accuracy performance on the leaderboards.



 In this work, we propose to evaluate models' understanding and consistency in predictions across modalities.  
 For that intention, we propose a Multimodal Evaluation (ME) evaluation schema that can augment existing VL benchmarks like VCR. For any given sample of the VL data, \textit{e.g.} image-question-answer pair, ME first retrieves and extracts related information of three modalities: vision, text, and background knowledge. After that, it unifies the information across modalities via a multimodal graph and further automatically generates related sub-questions corresponding to all three modalities (as examples shown in Fig. \ref{fig:motivation}). 
 
 The sub-questions would be semantically relevant to the input image-question-answer pair, and therefore, after answering the original input question, we can further utilize them to evaluate existing VL models' understanding across the three modalities and pinpoint their shortcoming and biases. Under minimal human verification, with ME, we create Multimodal Sub-question Evaluation Benchmark with 630k multiple choice sub-questions for 110k images from VCR \cite{vcr}: 110k of them are visual; 260k of them are about text; and the rest 260k are related to background knowledge.
 
 After in-depth evaluation and analysis with top-performing VL models, we discover a few interesting findings: (1) semantically low-level information can assist learning of high-level information but not the opposite; (2) visual information is generally under utilization compared with text. (3) VL models may struggle to utilize related background knowledge information.

Besides, we propose a Multimodal Coaching (MC) framework to conditionally augment sub-questions in training. Depending on VL models' behavior, MC would conditionally decide if it should augment to reinforce the understanding of a particular modality. We show that by using MC, we not only improve models' consistency but also the overall performance. For example, MC boosts the performance of VL-BERT by more than $1\%$ on the original VCR Q2A metric and even more then $7\%$ in sub-question evaluation metric.

Our contributions include:
\begin{enumerate}
  \setlength{\itemsep}{0pt}
  \setlength{\parskip}{0pt}

    \item We identify while existing VL models perform well in commonsense benchmark, they cannot answer related sub-questions.
    \item 
    Our proposed fine-grained automatic evaluation approach 
    allows the communities to better evaluate VL models. The code/dataset will be released upon acceptance.
    
    
    \item Our in-depth evaluation and analysis with top-performing VL models discover that:
    (1) Training with semantically low-level information may assist learning high-level concept but not the opposite; (2) Visual information is generally under utilized compared to textual information. 

\end{enumerate}

\vspace{-3mm}

\section{Related Work}



Biases occur if VL models cannot comprehensively understand the contents of both images and texts. They need to not only understand information from the two modalities respectively but also integrate these information by cross-referencing. \cite{cao2020behind, manjunatha2019explicit} pointed out biases like unbalanced utilization between visual and textual information. Based on these findings, previous works proposed different methods for countering them in VL benchmarks. For instance, \cite{agrawal2018don, zhang2016yin, dancette2021beyond} diversifies and shifts VQA's answer distribution \cite{goyal2017making} to balance the dataset; \cite{gokhale2020mutant, liang2020learning, gupta2022swapmix, liang-etal-2020-learning} augments images or creates counterfactual images to train more robust models on VQA; \cite{niu2021introspective, ramakrishnan2018overcoming, niu2021counterfactual, wang2022multimodal, zhang2021multi} regularizes models' training with prior knowledge to avoid learning biases; \cite{ye2021case} directly aligns pronouns to demonstrate biases in VCR \cite{vcr}\textit{, etc.} However, none of them helps us evaluate VL models' understanding on each modality. Without understanding how much models understand the image, the text, or the background knowledge, it is difficult to further regularize models in training. 

 Recently, large pretrained VL models, which are mostly trained as implicit black boxes, have been dominating VL benchmarks. It is difficult to know if they understand the image and the textual information other than simply memorizing it. Question-answering is the most general format for evaluating a wide range of models while having minimal requirements. \cite{ray2019sunny,ribeiro-etal-2019-red, selvaraju2020squinting} annotated additional questions on top of VQA questions to measure VL models' consistency on prediction. However, these works only focus on semantically low-level dataset like VQA and do not apply to highly semantic dataset like VCR. Moreover, their data fully relies on manual annotation and thus is hard to scale. Furthermore, they also fail to evaluate models' understanding across modalities. To solve these problems, we create a VL evaluation method that generates data with minimal human efforts, differentiates evaluation between modalities, and applies to highly semantic VL benchmarks. \footnote{This paper mainly focuses on applying ME on VCR, but our method can also be applied to other VL dataset consisted by image-text pairs. We also have tried it on Visual Question Answering (VQA) \cite{li2018vqa} and Visual Entailment (SNLI-VE) dataset \cite{xie2018visual}. Details in Appendix.}

\vspace{-2mm}

\begin{figure*}[htpb!]
\begin{center}
\scalebox{1}{
  \includegraphics[width=\linewidth]{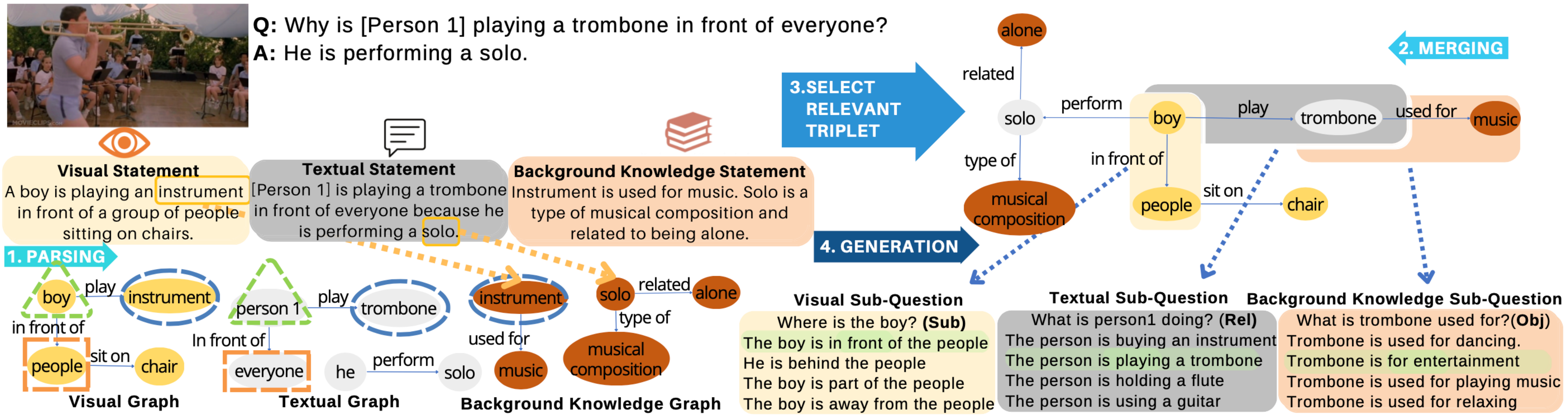}
}
\end{center}
  \caption{Pipeline of Multimodal Question-Answer Generation.}
\label{fig:pipeline}
\vspace{-6mm}
\end{figure*}

\section{Multimodal-Eval (ME)}
 Given an input image-question(-answer) pair, ME first analyzes information from the pair. Then it would generate three follow-up fine-grained questions called sub-questions corresponding to three modalities: vision, text, and background knowledge. Following VCR's format, each sub-question also has four answer choices with one correct answer. The VL models are expected to first answer the VCR question and then answer the three related sub-questions. Through evaluating predictions of the sub-questions, we can test models' understanding across modalities. Overall, ME has two parts: (1) \textbf{Multimodal QA Generator} that generates related sub-questions of three modalities and (2) \textbf{Evaluation}, which test VL models' capabilities with the sub-questions. We structure the presentation as below: the method section explains the QA generation process and the experiment section discusses the evaluation process.

\section{Multimodal QA Generator}

We introduce Multimodal QA Generator through the following steps in order: (1) Retrieving related sentence statements of three modalities against the input image-text, (2) Parsing the statements into three unimodal graphs and then merging them into a multimodal graph, (3) Converting triplets in a multimodal graph into question and answer, (4) Distractor Generation, (5) Adversarial Filtering.






\subsection{Retrieving Statements}
For producing relevant sub-questions, we need to first analyze the input image-text pair and even extract information from it. Therefore, it is intuitive to have the input image and text information represented in the same level of complexity. Because the input question-answer is already in text format, we want to convert the image into text. 

\noindent\textbf{Visual Statement: }Most of the existing highly-semantic VL benchmarks build on top of image/video captioning dataset, \textit{e.g.} VQA \cite{goyal2017making} from COCO Captions \cite{chen2015microsoft}, SNLI-VE \cite{xie2018visual} from Flickr30K \cite{young-etal-2014-image}, VCR from LSMDC \cite{rohrbach2017movie}\textit{, etc.} Those captions are visually descriptive and are not included in the image-question-answer pair. Therefore, we can directly retrieve those already annotated captions as visual statement. 

\noindent\textbf{Textual Statement: }The input text prompt, \textit{e.g.} the question-answer pairs in VCR, can be converted into statements with heuristic templates. For instance, the QA in Fig. \ref{fig:pipeline} can be converted to "Person1 plays a trombone in front of everyone" + "because" + "he is performing a solo". We can also regard the converted statement as textual statement, as shown in Fig. 2 (Details in Appendix). 

\noindent\textbf{Background Knowledge Statement: }In order to obtain background knowledge relevant to the visual scene, we apply keyword extractors \cite{yake} to extract keywords from visual and textual statements. Then we can regard those keywords as query concepts (as illustrated in Fig. \ref{fig:pipeline}, query concepts "trombone" and "solo" are extracted from the visual and textual statements). Based on query concepts, we can further browse external knowledge database, \textit{i.e.} ConceptNet \cite{speer2017conceptnet}
to retrieve 1-hop related concepts\footnote{https://github.com/ldtoolkit/conceptnet-lite}
through a pool of hand-selected relationships\footnote{PartOf, IsA, HasSubevent, Synonym, Antonym, MadeOf, DerivedFrom, DefinedAs, RelatedTo, UsedFor, CapableOf, AtLocation, Causes, HasProperty, Desires, CreatedBy, DistinctFrom, SymbolOf, LocatedNear, SimilarTo}. As illustrated in background knowledge graph in Fig. \ref{fig:pipeline}, different triplets consisted of (Subject, Predicate, Object) are retrieved and can be conveniently converted into basic Subject-Verb-Object (SVO) sentences. 



\subsection{Generating Graph}

For better integrating information, we leverage a language parser to parse the statements so that we can obtain semantic roles: Subject, Predicate and Object (S, P, O). These roles help further unify fine-grained information across three modalities. With comparing them, we may find connections.

 \noindent\textbf{Domain-specific Graph: }The background knowledge we retrieved from ConceptNet is already in a graph consisted of triplets (S, P, O). 
Therefore, we only apply the Scene Graph Parser \cite{schuster2015generating} to parse visual and textual statements into graphs. As shown in Fig. 2, we then have three domain-specific graphs corresponding to all three modalities.


\noindent\textbf{Multimodal Graphs: }
To merge two graphs, 
we take turns to compare the similarity between each pair of nodes from them. During comparison, we not only measure their concepts' similarity but also their neighbors/context similarity. If they are similar, they would be merged into one node.
For instance, given graph $G_1$ and $G_2$, we compare every node $v_{i}, i \in[0, \ldots, n]$ in $G_1$ with every node $v_{j}, j \in[0, \ldots, m]$ in $G_2$. We calculate the semantic similarity score between them, $sim_{c}\left(v_{i}, v_{j}\right)$ through an external tool \cite{zhu2017sematch}\footnote{It measures the distance of the two nodes' concepts in WordNet \cite{wordnet} and YAGO \cite{Pellissier2020YAGO} and then averages the reverse of the two distances as the similarity score}.

Subsequently, we also compare the neighbors of $v_{i}$ against the neighbors of $v_{j}$. Let's assume $v_{i}$ has $p$ 1-hop connections in $G_1$ and $v_{j}$ has $q$ 1-hop connections in $G_2$. Every connection of $v_{i}$ links two concepts thus forming a triplet. It can be converted into a SVO sentence containing $v_{i}$ as either the Subject or Object. With that, we result in $p$ sentences related to $v_i$. Similarly, we can also obtain $q$ sentences related to $v_j$. Following \cite{ni-etal-2022-sentence}, we inference a pretrained Sentence-T5 to extract $p$ and $q$ sentence embeddings. Then, for every pair between $S_{l}, l \in[0, \ldots, p]$ and $S_{o}, o \in[0, \ldots, q]$, we calculate the cosine distance $sim_{s}\left(S_{l}, S_{o}\right)$. Lastly, for every pair, $(v_{i}, v_{j}$),  we sum both node concept similarity and context similarity together by:

\begin{equation}
\begin{aligned}
\operatorname{Score}_{node}\left(v_{i}, v_{j}\right)=\operatorname{sim}_{c}\left(v_{i}, v_{j}\right) + \\
\frac{\sum_{o} \sum_{l} \operatorname{sim}_{s}\left(S_{l}, S_{o}\right)}{p \cdot q}.
\end{aligned}
\end{equation}
 If $\operatorname{Score}_{node}\left(v_{i}, v_{j}\right)$ is larger than a threshold $T$ (Details in Appendix), we would consider $v_{i}, v_{j}$ as duplicates and only keep one in the graph.



\noindent\textbf{Selecting Relevant Sub-graphs: }After obtaining the graph representation, we want to generate sub-questions relevant to the input image and VCR question of each sample $u$. Therefore, we filter each triplet in the multimodal graph by its 
relevance against the input image-question-answer pair.

 Similar to above, we convert all $r$ triplets in multimodal graph into sentences $S_{k}^{u}, k \in[0, \ldots, r]$ and measure their similarity to the textual statement (a conversion of the input QA) via \cite{ni-etal-2022-sentence}, $sim_{s}\left(S_{k}^{u}, S_{QA}^{u}\right)$. 

 Afterwards, we further utilize a pretrained CLIP \cite{radford2021learning} to encode and then calculate the cosine distance between every sentence against the image $I^{u}$, $rel_{s}\left(S_{k}^{u}, I^{u}\right)$. In conclusion, the final score for every triplet would be:
\begin{equation}
\left\|\operatorname{sim}_{s}\left(S_{k}^{u}, S_{Q A}^{u}\right)||_{1}+\right\| rel_{s}\left(S_{k}^{u}, I_{u}\right)||_{1}.
\end{equation}

 After ranking, we select the top-1 ranked triplet in every modality. If a triplet is selected in more than one modality, we replace the duplicate with the next following triplet in the same modality.

\subsection{QA Templates}
 Given a triplet, we can ask questions about the subject, the object, or the predicate. For instance, in (boy, in front of, people), if asking about the object, we could use templates like "What is the [Subject] [Predicate]" (What is the boy in front of?). In this case, the basic answer would be [Object](People) or the converted full SVO sentence of the triplet, [Subject][Predicate][Object](The boy is in front of people). When asking about the subject or the predicate, similar procedure applies (Details in Appendix)


\subsection{Distractor Generation}
 The new evaluation task should have the same format as the original one (multiple-choice-format (MCQ) in VCR) so that we can directly evaluate existing models. For that purpose, it is necessary to generate incorrect answer choices, distractors.

Simply rephrasing the correct answer may produce false negative that confuse the models. In a sense, more non-trivial and meaningful disturbance should be added to the answer distribution. 

 In practice, we choose to represent the answer in SVO sentence format,  \textit{e.g.} ``The boy is in front of people''. We first parse the answer into (Subject, Predicate, Object) \textit{e.g.} (boy, in front of, people) and regard this as the starting templates for creating distractors.
If the question is asking about the relationship, then we could regard relationship as the ``changeable part'' in the template. We could replace this ``changeable part'' with other words to create new combinations for distractors \textit{e.g.} (boy, behind, people). In order to make meaningful replacement, we use the original relationship concept, ``in front of'', as the query concepts to retrieve related concepts from external resources like ``behind'', ``direction'', ``location''. We apply the same procedure to the subject and object.


 \noindent\textbf{Explicit Retrieval from External Knowledge: }
 We follow a similar procedure in retrieving background knowledge concepts from ConceptNet \cite{speer2017conceptnet}, while only differing in our selection of a different set of relationships 
(Details in the Supple). 

 \noindent\textbf{Implicit Retrieval from Language Models: }
We also utilize pretrained language models to help retrieve related concepts in two perspectives.
First, in cases when the question is asking about the object and the program fails to retrieve related concepts from explicit resources, we leverage prompt engineering to implicitly retrieve related concepts from a pretrained language model, GPT2 \cite{radford2019language} alternatively. Using the same triplet as an example, if the question asks about the object, then we would design the prompt as ``boy is in front of [mask]''. After GPT2 fills in the [MASK], we should be able to retrieve external concepts within GPT2's top predictions. We can further use as them as options for objects in distractors. 

After successfully replacing concepts in the template, we directly apply heuristic rules and convert it into SVO sentences with heuristic rules to create a distractor. Aiming for variety beyond rule-based sentence construction, we also alternatively use another language model to process the conversion.
We exploit a sentence generation model, T5 finetuned on CommonGen \cite{lin-etal-2020-commongen} 
which is built on top of ConceptNet. The training task in CommonGen is to convert set of concpet words into everyday sentences. 
For example, after replacing concepts in the template, from \textit{(boy, in front of, people)} to \textit{(boy, back, people)} and \textit{(boy, direction, people)}, we input them directly into T5, which outputs a list of possible sentences \textit{e.g.} ``A boy is running back to the people'', ``A boy is facing the same direction as the other people''. Different from hard-coded templates used to generate SVO sentences, T5 fills in context words around the input concepts, thus also helps retrieving implicit external concepts like (``running'', ``facing'', ``same'', ``other''). These additional concepts may not be relevant to the visual scene which aligns with the purpose of generating distractors.



\subsection{Adversarial Filtering}
 High-quality distractors should be semantically related to the answer but also different enough for humans to tell. Therefore, we design our own adversarial filtering \cite{zellers2018swag, vcr} mechanism by using pretrained VL and language models to filter data. We first correct all generated distractors by an off-shelf grammar checker\footnote{https://pypi.org/project/language-tool-python}. Then we further filter them by a pretrained language model to remove distractors that are too semantically close to the correct answer to reduce potential false negatives. Lastly, we apply a pretrained VL model to measure their relevance against the image and select the top three as final distractors (Details in Appendix).





\vspace{-2mm}

\section{Dataset}
\paragraph{Dataset Statistics}

Built on top of \cite{vcr}, Multimodal Sub-question Evaluation Benchmark has around 110k visual sub-questions corresponding to the 110k images from \cite{vcr}, 260k text(prompt) sub-questions, and 260k background knowledge sub-questions corresponding to the 290k original questions from \cite{vcr}\footnote{One image has only one visual sub-question but may correspond to multiple
text or background knowledge sub-questions. Some of the original VCR questions are too short and do not contain meaningful sub-questions}. Every question has four answer choices and the answers have an average length of 5.5 words. The ratio between training set and validation set is $10:1$.

\paragraph{Quality Control}
To deliver a convincing evaluation method to existing VL models, we have humans verify the full validation set. We designed and deployed a user interface on Amazon Mechanical Turk platform and hired experienced turkers (with $\$12.6/hr$)to help verify the correctness of our questions and answers. Every image-question pair was cross-verified and corrected by five turkers (Details in Appendix).

 \begin{table}[t]
\centering
\resizebox{6cm}{!}{%
\centering
\begin{tabular}{|c|c|c|}
\hline
Metric           & Generated & Verified \\ \hline
Individual Acc.  & 0.83      & 0.89     \\ \hline
Group Acc.       & 0.95      & 0.99     \\ \hline
Group Top2 Recall & 0.94      & 0.98     \\ \hline
IAA              & 0.82      & 0.88     \\ \hline
\end{tabular}
}
\caption{Comparison between generated and verified data. Every sample has five annotations/selections. Individual Acc. represents the accuracy when each annotator's selection is counted as one prediction. Group Acc. represents the accuracy when only the highest frequent selection is counted as the prediction for the group. Group Top2 Recall represents the accuracy if the groundtruth is within the top 2 most frequent selections of the group. IAA is the Inter-Annotator Agreement}
\vspace{-6mm}
\label{annotation}
 \end{table}


\paragraph{Evaluation}

We randomly select 2 disjoint sets each containing 100 image-question pairs from ME. The first set consists generated QA data. The second one consists QA data verified and corrected by the turkers. We then hire an additional group of five turkers to answer those 200 image-question pairs without knowing the answer label. Next, we calculate the predictions' accuracy. As in Tab. \ref{annotation}, the difference between generated and verified data is very minimal which demonstrates the high-quality of our generated data (More in Appendix).



\vspace{-3mm}

\section{Evaluation}

In the following, we conduct experiments based on the proposed dataset to demonstrate (1) The existing models that perform well on VL dataset often cannot answer detailed vision, text, knowledge sub-question correctly; (2) It is easier for VL models to answer sub-questions originated from semantically low-level VCR questions than high-level ones.

\paragraph{Base Methods}
During our experiments, we use three top-performing models, VL-BERT \cite{su2019vl}, UNITER \cite{chen2020uniter} and VILLA \cite{gan2020large} on VCR leaderboard as our base models.

\paragraph{Evaluation Metrics}
When calculating the original Q2A accuracy of VCR \cite{vcr} on $n$ total samples, let ${C}_{j}^{Q2A}$ be an indicator variable for sample $j, j \in[0, \ldots, n]$. If the prediction, $P_{j}^{Q2A}$, is the same as the label, $L_{j}^{Q2A}$, the prediction is correct and ${C}_{j}^{Q2A}=1$; otherwise ${C}_{j}^{Q2A}=0$.

\vspace{-6mm}


$$
  \text { Accu. }_{\mathrm{Q} 2 \mathrm{A}}=\frac{\sum_{j=0}^{j=n} \operatorname{C}_{j}^{Q2A}}{n}.
$$

Similarly, in our new metrics, we have indicator variables $\text {C}_{j}^{Q2S-x}$ for the correctness of prediction on sub-questions related to modality $x$, which can be vision, text, or background knowledge; similarly, we use $\text {C}_{j}^{Q2AS-x}$ to indicate the event that both the VCR question and the sub-question corresponding to modality $x$ are correct. Lastly, $\text{C}_{j}^{Q2S}$ indicates the event that all the sub-questions related to sample $j$ are predicted correctly.
$$
\left\{\begin{array}{l}

\text {C}_{j}^{Q2S-x}=1 \text {, if } P_{j}^{Q2S-x}=L_{j}^{Q2S-x}, \text { else } 0 \\
\text {C}_{j}^{Q2AS-x}=1 \text {, if C}_{j}^{Q2A}=1 \text { and }
\text {C}_{j}^{Q2S-x}=1 \\
\text {C}_{j}^{Q2S}=1 \text {, if } \sum \text { C}_{j}^{Q2S-x}=3, x \in\{V, T, B K\}
\end{array}\right.
$$



\subsection{Comparison across Modalities}
We want to evaluate VL models' capability in understanding fine-grained information from different modalities. Tab. \ref{consistency} shows the evaluation results of the models. Looking at rows marked with "N" under the column name "ME in Training", we discover that existing VL models all suffer around a $20\%$ drop in accuracy on our sub-questions' metrics. Among modalities, VL models generally perform the best in textual sub-questions. This is expected since the semantic contents of the textual sub-questions are the closest one to the original VCR questions. In contrast, these models often perform slightly worse in visual sub-questions. This re-verifies previous works' concerns that existing VL models generally under-utilize visual information. Lastly, they all suffer the most in answering background knowledge sub-questions. We believe that despite background knowledge may be useful in humans' perspective, VL models are still lack of sufficient abilities to utilize them. In fact, they seem to have the largest domain gap against the original VCR questions to VL models. These evaluation results help verify our previous hypothesis.

 \noindent\textbf{Consistency: }When considering consistency, even larger drops about $40\%$ occur across models' performances. The general trend of performances on Q2AS-x between modalities is similar to Q2S-x as discussed before, but with lower overall values.

\begin{table*}[!htbp]
\centering
\resizebox{12cm}{!}{%
\begin{tabular}{|c|c|ccccclll|}
\hline
\textbf{Model}              & \textbf{\begin{tabular}[c]{@{}c@{}}ME \\ in Training\end{tabular}} & \multicolumn{8}{c|}{\textbf{Evaluation}}                                                                                                                                                                                                                                                                                                    \\ \hline
                            &                                                                    & \multicolumn{1}{c|}{\textbf{VCR}} & \multicolumn{4}{c|}{\textbf{Subsequent Questions}}                                                                                                                   & \multicolumn{3}{c|}{\textbf{Consistency}}                                                                                        \\ \cline{3-10} 
\multirow{-2}{*}{\textbf{}} & \multirow{-2}{*}{\textbf{}}                                        & \multicolumn{1}{c|}{\textbf{Q2A}} & \multicolumn{1}{c|}{\textbf{Q2S}}                 & \multicolumn{1}{c|}{\textbf{Q2S-V}} & \multicolumn{1}{c|}{\textbf{Q2S-T}} & \multicolumn{1}{c|}{\textbf{Q2S-BK}} & \multicolumn{1}{c|}{\textbf{Q2AS-V}}              & \multicolumn{1}{c|}{\textbf{Q2AS-T}} & \multicolumn{1}{c|}{\textbf{Q2AS-BK}} \\ \hline
                            & N                                                                  & \multicolumn{1}{c|}{75.53}        & \multicolumn{1}{c|}{{\color[HTML]{333333} 55.31}} & \multicolumn{1}{c|}{54.96}          & \multicolumn{1}{c|}{56.18}          & \multicolumn{1}{c|}{55.75}           & \multicolumn{1}{l|}{{\color[HTML]{333333} 41.59}} & \multicolumn{1}{l|}{42.51}           & 42.19                                 \\ \cline{2-10} 
\multirow{-2}{*}{VL-BERT}   & Y                                                                  & \multicolumn{1}{c|}{76.59}        & \multicolumn{1}{c|}{61.16}                        & \multicolumn{1}{c|}{60.12}          & \multicolumn{1}{c|}{62.81}          & \multicolumn{1}{c|}{58.75}           & \multicolumn{1}{l|}{46.05 (+4.46)}                & \multicolumn{1}{l|}{48.11 (+5.6)}    & 44.99 (+2.8)                          \\ \hline
                            & N                                                                  & \multicolumn{1}{c|}{76.64}        & \multicolumn{1}{c|}{{\color[HTML]{333333} 57.49}} & \multicolumn{1}{c|}{57.83}          & \multicolumn{1}{c|}{57.54}          & \multicolumn{1}{c|}{56.34}           & \multicolumn{1}{l|}{{\color[HTML]{333333} 44.32}} & \multicolumn{1}{l|}{44.1}            & 43.17                                 \\ \cline{2-10} 
\multirow{-2}{*}{UNITER}    & Y                                                                  & \multicolumn{1}{c|}{77.12}        & \multicolumn{1}{c|}{63.51}                        & \multicolumn{1}{c|}{62.84}          & \multicolumn{1}{c|}{66.04}          & \multicolumn{1}{c|}{60.87}           & \multicolumn{1}{l|}{48.46 (+4.14)}                & \multicolumn{1}{l|}{50.93 (+6.83)}   & 46.94 (+3.77)                         \\ \hline
                            & N                                                                  & \multicolumn{1}{c|}{78.27}        & \multicolumn{1}{c|}{{\color[HTML]{333333} 59.85}} & \multicolumn{1}{c|}{58.41}          & \multicolumn{1}{c|}{61.05}          & \multicolumn{1}{c|}{56.55}           & \multicolumn{1}{l|}{{\color[HTML]{333333} 45.71}} & \multicolumn{1}{l|}{47.78}           & 44.26                                 \\ \cline{2-10} 
\multirow{-2}{*}{VILLA}     & Y                                                                  & \multicolumn{1}{c|}{78.79}        & \multicolumn{1}{c|}{63.99}                        & \multicolumn{1}{c|}{63.2}           & \multicolumn{1}{c|}{66.43}          & \multicolumn{1}{c|}{60.83}           & \multicolumn{1}{l|}{48.74 (+3.03)}                & \multicolumn{1}{l|}{51.23 (+3.45)}   & 46.91 (+2.65)                         \\ \hline
\end{tabular}%
}
\caption{Evaluation of benchmark VL models' consistency across modalities.}
\label{consistency}
 \end{table*}

\vspace{-2mm}

\subsection{Comparison across Question Types}

\begin{figure*}[ht!]
\begin{center}
\scalebox{0.8}{
  \includegraphics[width=\linewidth]{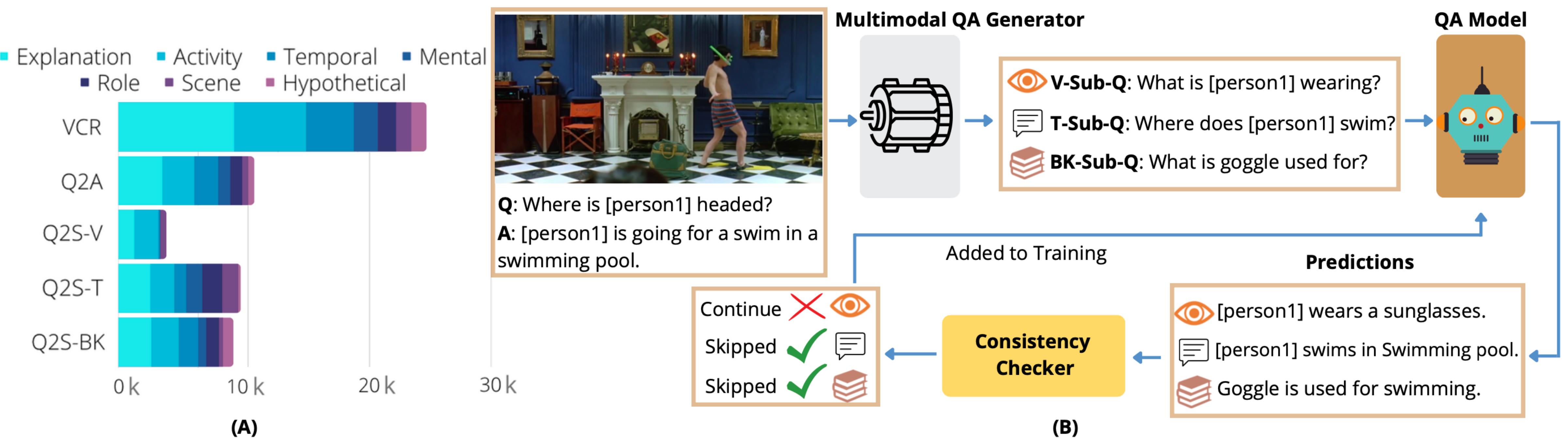}
}
\end{center}
  \vspace{-3mm}
  \caption{(A) Evaluation across Question Types. (B) Pipeline of Multimodal Coaching.}
\label{fig:chart}
\end{figure*}

The first row in Fig. \ref{fig:chart} (A) visualizes the number of questions of each type in VCR validation set. Observing it, we can see clear imbalanced distribution exists among question types in VCR validation set. Hence, we on purpose, sample 2k questions of each type from validation set to create a balanced mini-validation set of 14k VCR image-question pairs. We further evaluate the finetuned VL-BERT on this mini-validation set. On the second row, for each question type, we visualize the number of Q2A questions that VL-BERT predicts correctly. From the third to the fifth row, we visualize sub-questions (originated from different types of Q2A questions) that VL-BERT predicts correctly. As we can see, for Q2S-V, explanation, activity and scene questions have the most percentage. Apart from the explanation and activity questions also being the most dominant question types in the training set, the semantic relatedness with the activity questions and the explanation questions also helps models answer visual sub-questions. Additionally, we realize that it is easier for the model to answer sub-questions (from all three modalities) of semantically low-level VCR questions like explanation, activity types. However it is difficult for abstract ones like mental questions.


\vspace{-3mm}

\section{Multimodal Coaching for Model Improvement}

Besides utilizing ME data for evaluating existing VL models' performance of fine-grained understanding across modalities and prediction consistency, we also find that ME can further assist existing VL models' training.

\vspace{-2mm}

\subsection{Multimodal Coaching}
In order to better utilize ME data and allow VL models to have a balanced learning over information from different modalities, following \cite{ray2019sunny}, we design the Multimodal Coaching (MC) system. According to (B) in Fig. \ref{fig:chart}, when iterating over every VCR sample in training, Multimodal QA generator produces relevant sub-questions. Then MC would take turns to test the QA model with those sub-questions across three modalities. If the model fails on any of them, the corresponding sub-question would be added to the training pool otherwise it would be passed. Therefore, we selectively augment ME data with VCR data during the training.

\vspace{-2mm}

\subsection{Data Augmentation}

 \begin{table*}[!htbp]
\centering
\resizebox{15cm}{!}{%
\begin{tabular}{|lllll|llllll|}
\hline
\multicolumn{5}{|c|}{{\color[HTML]{333333} \textbf{Training}}}                                                                                                                                                                                                  & \multicolumn{6}{c|}{{\color[HTML]{333333} \textbf{Evaluation}}}                                                                                                                                                                                                                                                         \\ \hline
\multicolumn{1}{|c|}{{\color[HTML]{333333} \textbf{VCR}}} & \multicolumn{1}{c|}{{\color[HTML]{333333} \textbf{Sub-V}}} & \multicolumn{1}{c|}{{\color[HTML]{333333} \textbf{Sub-T}}} & \multicolumn{1}{c|}{{\color[HTML]{333333} \textbf{Sub-BK}}} & \textbf{MC} & \multicolumn{1}{c|}{{\color[HTML]{333333} \textbf{Q2A}}}             & \multicolumn{1}{l|}{\textbf{QA2R}}          & \multicolumn{1}{c|}{{\color[HTML]{333333} \textbf{Q2S}}}             & \multicolumn{1}{c|}{Q2S-V}                    & \multicolumn{1}{c|}{Q2S-T}                    & \multicolumn{1}{c|}{Q2S-BK} \\ \hline
\multicolumn{1}{|l|}{{\color[HTML]{333333} Y}}            & \multicolumn{1}{l|}{{\color[HTML]{333333} }}               & \multicolumn{1}{l|}{{\color[HTML]{333333} }}               & \multicolumn{1}{l|}{{\color[HTML]{333333} }}                &             & \multicolumn{1}{l|}{{\color[HTML]{333333} 75.67}}                    & \multicolumn{1}{l|}{77.84}                  & \multicolumn{1}{l|}{{\color[HTML]{333333} 55.31}}                    & \multicolumn{1}{l|}{54.96}                    & \multicolumn{1}{l|}{56.18}                    & 55.75                       \\ \hline
\multicolumn{1}{|l|}{{\color[HTML]{333333} Y}}            & \multicolumn{1}{l|}{{\color[HTML]{333333} Y}}              & \multicolumn{1}{l|}{{\color[HTML]{333333} }}               & \multicolumn{1}{l|}{{\color[HTML]{333333} }}                &             & \multicolumn{1}{l|}{{\color[HTML]{333333} 76.08   (+0.41)}}          & \multicolumn{1}{l|}{78.33 (+0.49)}          & \multicolumn{1}{l|}{{\color[HTML]{333333} 59.07   (+3.76)}}          & \multicolumn{1}{l|}{59.84   (+4.88)}          & \multicolumn{1}{l|}{59.51   (+4.33)}          & 58.01   (+2.26)             \\ \hline
\multicolumn{1}{|l|}{{\color[HTML]{333333} Y}}            & \multicolumn{1}{l|}{{\color[HTML]{333333} Y}}              & \multicolumn{1}{l|}{{\color[HTML]{333333} Y}}              & \multicolumn{1}{l|}{{\color[HTML]{333333} }}                &             & \multicolumn{1}{l|}{{\color[HTML]{333333} 76.59    (+0.92)}}         & \multicolumn{1}{l|}{78.76 (+0.92)}          & \multicolumn{1}{l|}{{\color[HTML]{333333} 61.16   (+5.85)}}          & \multicolumn{1}{l|}{60.12   (+5.16)}          & \multicolumn{1}{l|}{62.81   (+6.63)}          & 58.75   (+3.00)             \\ \hline
\multicolumn{1}{|l|}{Y}                                   & \multicolumn{1}{l|}{Y}                                     & \multicolumn{1}{l|}{Y}                                     & \multicolumn{1}{l|}{Y}                                      &             & \multicolumn{1}{l|}{76.48   (+0.81)}                                 & \multicolumn{1}{l|}{78.35 (+0.51)}          & \multicolumn{1}{l|}{59.14   (+3.83)}                                 & \multicolumn{1}{l|}{58.63   (+2.67)}          & \multicolumn{1}{l|}{60.66   (+4.48)}          & 59.47   (+3.72)             \\ \hline
\multicolumn{1}{|l|}{Y}                                   & \multicolumn{1}{l|}{Y}                                     & \multicolumn{1}{l|}{Y}                                     & \multicolumn{1}{l|}{}                                       & Y           & \multicolumn{1}{l|}{{\color[HTML]{333333} \textbf{76.88   (+1.21)}}} & \multicolumn{1}{l|}{\textbf{79.05 (+1.21)}} & \multicolumn{1}{l|}{{\color[HTML]{333333} \textbf{61.89   (+6.58)}}} & \multicolumn{1}{l|}{\textbf{60.44   (+5.48)}} & \multicolumn{1}{l|}{\textbf{63.62   (+7.44)}} & \textbf{59.41   (+3.66)}    \\ \hline
\end{tabular}%
}
\caption{Data augmentation. Numbers in brackets are the difference between data in that row against the first row.}
\label{ablation}
\vspace{-4mm}
 \end{table*}

We demonstrate the effectiveness of training with ME data  in Tab. \ref{ablation}. We keep VL-BERT as our base model and cumulatively add sub-questions across three modalities into the training set. VCR has 7 types of questions and some of them are highly semantic not much related to visual compositions like mental, hypothetical questions. With this prior knowledge, when adding visual sub-questions, we on purpose do not augment them to VCR questions of these two types.


We observe in Tab. \ref{ablation} that adding visual and textual sub-questions both bring improvements on Q2A, QA2R and sub-question metrics including Q2S, Q2S-V, Q2S-T and Q2S-BK. However, adding background knowledge sub-questions hurts the performance. As mentioned before in \textbf{Evaluation} section, the additional content from external database like ConceptNet has a large domain gap against VCR questions and thus may be too difficult for VL models to utilize. However, this result further debunks existing VL models' vulnerability and confirms that it is important to include background knowledge sub-questions in VL evaluation analysis.

Lastly, looking at the last row of Tab. \ref{ablation}, we observe that MC could further boost VL-BERT's performance gain. In experiments, we also realize that adding MC would allow training loss to be more stable and converge faster.

\subsection{Composite vs. Component Information}

 \begin{table}[!htbp]
\centering
\resizebox{7.5cm}{!}{%
\begin{tabular}{|cccc|ll|}
\hline
\multicolumn{4}{|c|}{\textbf{Training}}                                                                                          & \multicolumn{2}{c|}{\textbf{Evaluation}}                                    \\ \hline
\multicolumn{1}{|c|}{\textbf{VCR}} & \multicolumn{1}{c|}{\textbf{Sub-V}} & \multicolumn{1}{c|}{\textbf{Sub-T}} & \textbf{Sub-BK} & \multicolumn{1}{c|}{\textbf{VCR (Q2A)}} & \multicolumn{1}{c|}{\textbf{Q2S}} \\ \hline
\multicolumn{1}{|c|}{Y}            & \multicolumn{1}{c|}{}               & \multicolumn{1}{c|}{}               &                 & \multicolumn{1}{l|}{75.67}              & 55.31                             \\ \hline
\multicolumn{1}{|c|}{-}            & \multicolumn{1}{c|}{Y}             & \multicolumn{1}{c|}{}               &                 & \multicolumn{1}{l|}{59.31}              & 60.11 (+4.8)                      \\ \hline
\multicolumn{1}{|c|}{Y}            & \multicolumn{1}{c|}{Y}             & \multicolumn{1}{c|}{}               &                 & \multicolumn{1}{l|}{76.08 (+0.41)}      & 59.07 (+3.76)                     \\ \hline
\multicolumn{1}{|c|}{-}            & \multicolumn{1}{c|}{-}              & \multicolumn{1}{c|}{Y}              &                 & \multicolumn{1}{l|}{59.72}              & 61.01 (+5.70)                     \\ \hline
\multicolumn{1}{|c|}{Y}            & \multicolumn{1}{c|}{-}              & \multicolumn{1}{c|}{Y}              &                 & \multicolumn{1}{l|}{76.20 (+0.53)}      & 60.33 (+5.02)                     \\ \hline
\multicolumn{1}{|c|}{-}            & \multicolumn{1}{c|}{-}              & \multicolumn{1}{c|}{-}              & Y               & \multicolumn{1}{l|}{55.72}              & 58.99 (+3.68)                     \\ \hline
\multicolumn{1}{|c|}{Y}            & \multicolumn{1}{c|}{-}              & \multicolumn{1}{c|}{-}              & Y               & \multicolumn{1}{l|}{75.48 (-0.19)}      & 58.12 (+2.81)                     \\ \hline
\end{tabular}%

}
\caption{Comparison between training with composite and component data. Numbers in brackets are the difference between data in that row against the first row.}
\vspace{-6mm}
\label{composite}
 \end{table}

Comparing the first row against the third row in Tab. \ref{composite}, we notice that VL-BERT performs better on Q2A when having both VCR questions and visual sub-questions in training set. Comparing the second row against the third row, we also discover that VL-BERT performs better on Q2S when the training set only contains visual sub-questions. Adding VCR questions would actually hurt its performance on Q2S.

We observe similar results when comparing other sets of rows like the (first,  fourth, fifth) rows for text sub-questions, and the (first, sixth, seventh) rows for background knowledge sub-questions. Even though, when having both background knowledge sub-questions and VCR questions in training, the Q2A performance drops slightly (due to potential reasons explained above), the Q2S performance drops even much more due to adding VCR questions. Also, Q2A performance via training on background knowledge sub-questions only is even higher than the Q2S performance via training on VCR questions only(Both questions share the same MCQ format with four answer choices and random guess is $25\%$).

If we regard VCR questions as composite information since information from different modalities are combined together in the questions, we can then refer sub-questions as component information "parsed from" the composite information. Based on the comparison, we conclude that low-level component information could potentially help models' understanding of high-level composite information. However, after learning with high-level composite information, existing VL models may struggle to utilize the high-level to help understand low-level component information.




\subsection{Comparison across Modalities}

As in Tab. \ref{consistency}, after adding ME sub-question data in training, VL models generally improve in accuracy across Q2A, sub-question metrics and consistency metrics. Complementary to the findings in the \textbf{Evaluation} section, we discover that \textbf{(1)} VL models tend to have more consistent predictions in answering textual sub-questions; \textbf{(2)} Adding textual sub-questions in training also brings more improvements on sub-questions metrics corresponding to the other two modalities.

\vspace{-3mm}

\section{Conclusion}
In this work, we propose ME to thoroughly probe VL models' understanding across and between modalities. Our analysis brings new insights and our experiments show that ME boosts models' performance when used in training.

\section{Limitation}

ME requires the given image to have paired captions so they can be easily converted into visual statements. When absent, we can inference from a pretrained caption generator at the expense of accuracy. However, sometimes the visual caption generator may not fully captures the most salient activities in the image and thus produces trivial captions with limited contents. Therefore, it would be difficult for ME to extract related information from the caption to further create the visual sub-question.

Also, technically, for any VL dataset with image-question-answer pairs, ME should be able to generate sub-questions from three modalities. However, if the input question is very simple and focuses on semantically low-level information. It woud be challenge for ME to further extract and create sub-questions from three modalities. 

This study is solely based on English data and leverages linguistic structures in English so it cannot generalize to other languages.

\section{Appendix}

 \begin{table*}[!htbp]
\centering
\centering
\begin{tabular}{|c|ccccc|}
\hline
\textbf{Verified} & \multicolumn{5}{c|}{\textbf{Evaluation}}                                                                                                                                  \\ \hline
                  & \multicolumn{1}{c|}{\textbf{VCR (Q2A)}} & \multicolumn{1}{c|}{\textbf{Q2S}} & \multicolumn{1}{c|}{\textbf{Q2S-V}} & \multicolumn{1}{c|}{\textbf{Q2S-T}} & \textbf{Q2S-BK} \\ \hline
N                 & \multicolumn{1}{c|}{75.67}              & \multicolumn{1}{c|}{54.23}        & \multicolumn{1}{c|}{54.81}          & \multicolumn{1}{c|}{54.95}          & 54.87           \\ \hline
Y                 & \multicolumn{1}{c|}{75.67}              & \multicolumn{1}{c|}{55.31}        & \multicolumn{1}{c|}{54.96}          & \multicolumn{1}{c|}{56.18}          & 55.75           \\ \hline
\end{tabular}%
\caption{Evaluation with generated and verified data.}
\label{eval}
 \end{table*}

 \begin{table*}[!htbp]
\centering

\begin{tabular}{|c|ccccc|}
\hline
\textbf{Verified in Training} & \multicolumn{5}{c|}{\textbf{Evaluation}}                                                                                                                                  \\ \hline
                              & \multicolumn{1}{c|}{\textbf{VCR (Q2A)}} & \multicolumn{1}{c|}{\textbf{Q2S}} & \multicolumn{1}{c|}{\textbf{Q2S-V}} & \multicolumn{1}{c|}{\textbf{Q2S-T}} & \textbf{Q2S-BK} \\ \hline
N                             & \multicolumn{1}{c|}{76.13}              & \multicolumn{1}{c|}{59.34}        & \multicolumn{1}{c|}{60.77}          & \multicolumn{1}{c|}{61.74}          & 58.8            \\ \hline
Y                             & \multicolumn{1}{c|}{76.59}              & \multicolumn{1}{c|}{61.16}        & \multicolumn{1}{c|}{60.12}          & \multicolumn{1}{c|}{62.81}          & 58.75           \\ \hline
\end{tabular}%
\caption{Evaluation of VL-BERT trained with generated and verified ME data augmentation.}
\label{train}
 \end{table*}

\subsection{Generated vs. Verified}

In Tab. \ref{eval}, we evaluate VL-BERT with both generated ME data and data verified by human annotators. 

In Tab. \ref{train}, we finetune a VL-BERT with both generated and verified data by humans.

Results from both tables demonstrate the high-quality of our generated data.

\subsection{Hyper-perameter}
1. In practice, the semantic similarity between concepts of two nodes would be first standardized via z-score and then compared against a hyper-parameter $T$ of 0.8.

\subsection{Examples in other VL Benchmarks}

\begin{figure*}[ht!]
\begin{center}
  \includegraphics[width=\linewidth]{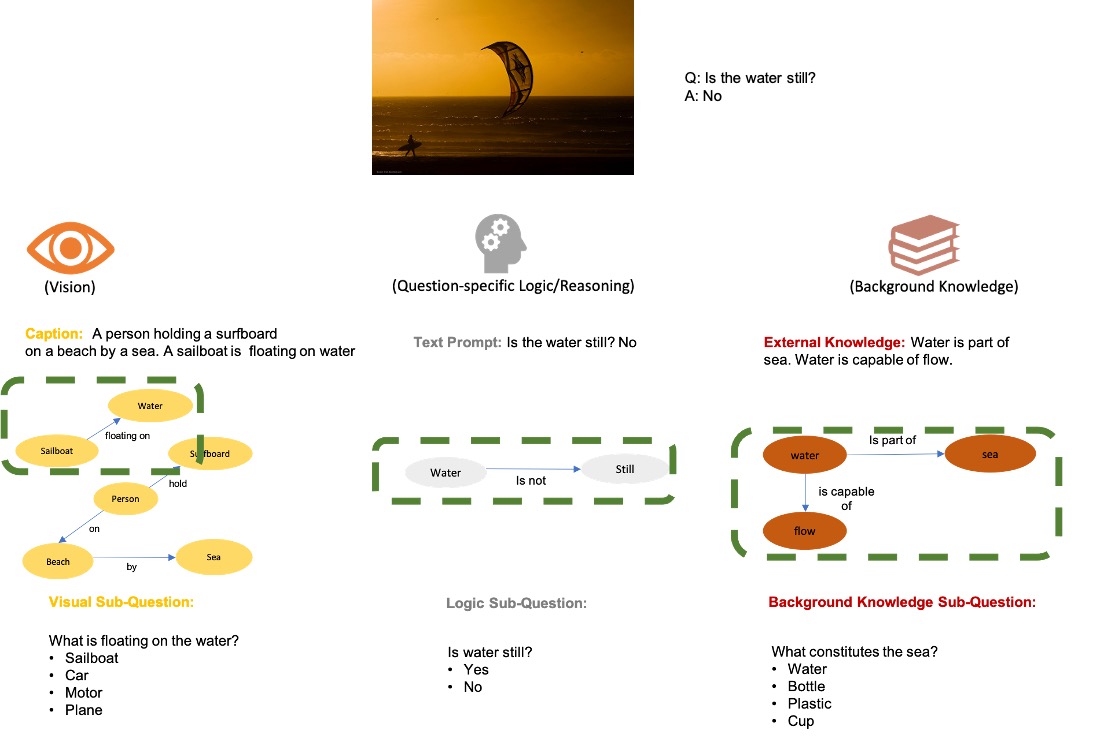}
\end{center}
  \caption{}
\label{fig:vqa}
\end{figure*}

\begin{figure*}[ht!]
\begin{center}
  \includegraphics[width=\linewidth]{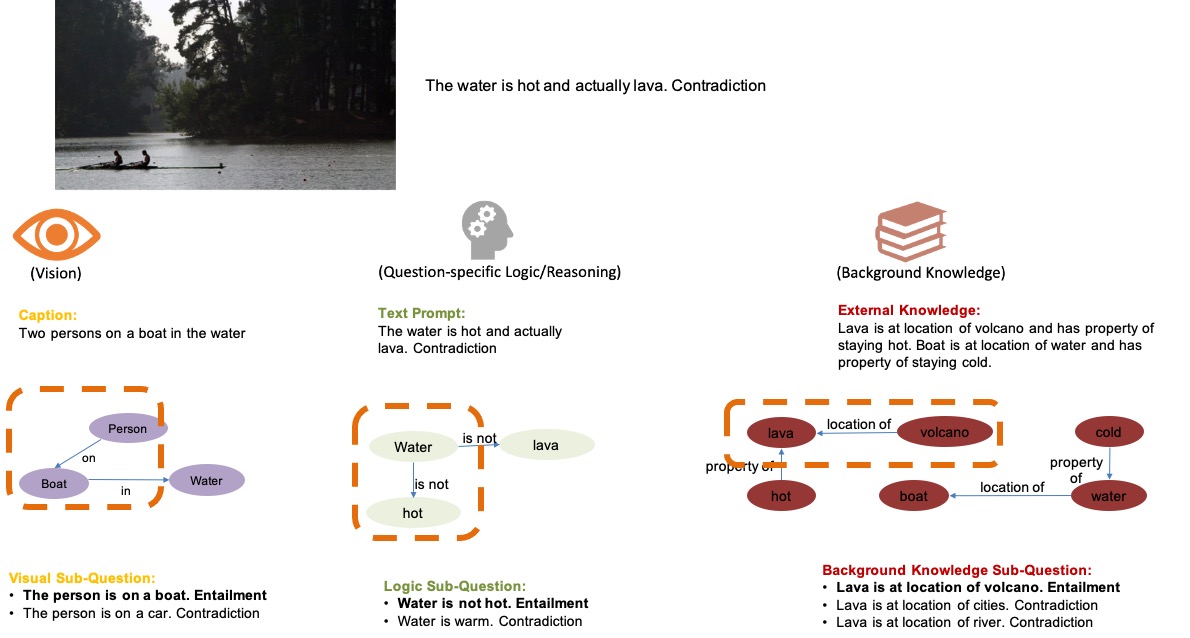}
\end{center}
  \caption{}
\label{fig:snlive}
\end{figure*}

Referring to Fig. \ref{fig:vqa}, \ref{fig:snlive}.

\subsection{User Interface}

\begin{figure*}[ht!]
\begin{center}
  \includegraphics[width=\linewidth]{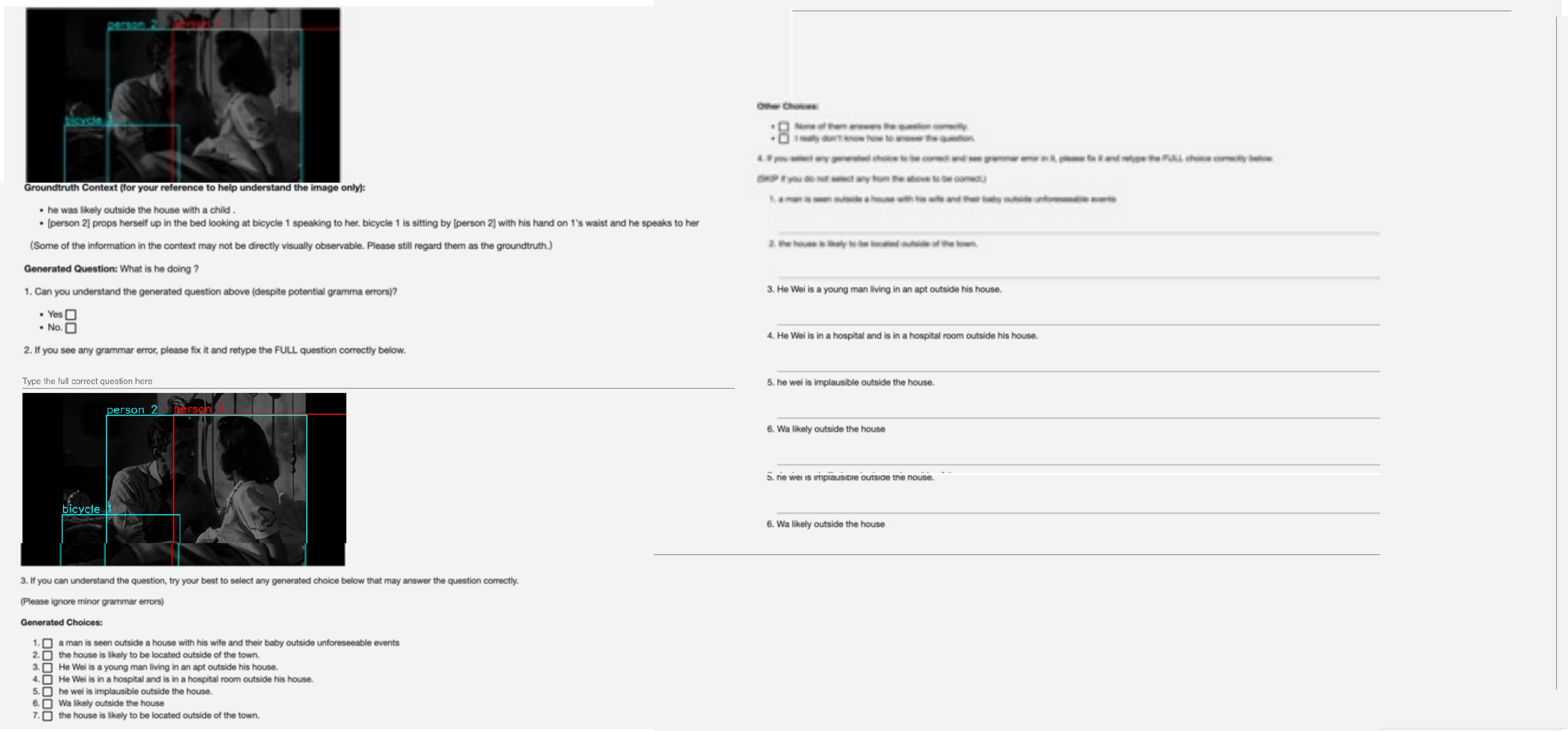}
\end{center}
  \caption{A user interface for collecting data.}
\label{fig:user interface}
\end{figure*}

Referring to Fig. \ref{fig:user interface}

\subsection{Adversarial Filtering}

High-quality distractors should be semantically related to the answer but also different enough for humans to tell. Therefore, we design our own adversarial filtering \cite{zellers2018swag, vcr} mechanism by using pretrained VL and language models to filter data. We first correct all generated distractors by an off-shelf grammar checker\footnote{https://pypi.org/project/language-tool-python}. Then we further filter them by a pretrained language model to remove distractors that are too semantically close to the correct answer to reduce potential false negatives. Lastly, we apply a pretrained VL model to measure their relevance against the image and select the top three as final distractors.

 \noindent\textbf{Sentence-Similarity Modeling: }
Similar to previous procedures, across all $z$ number of distractors, we compare each of them $S_{w}, w \in[0, \ldots, z]$ against the textual (QA) statement $S_{QA}^{u}$, $Score_{sent}^{w}=\operatorname{sim}_{s}\left(S_{w}^{u}, S_{Q A}^{u}\right)$. By removing distractors whose $Score_{sent}^{w}$ is above a threshold, $D$ (0.7), we reduce potential false negatives that are semantically close to the correct answer.



 \noindent\textbf{Image-Text Matching: }
After that, we also need to ensure that the distractors are visually relevant to the image. We load a pretrained CLIP model \cite{radford2021learning} to measure the relevance between each distractor against the image, $Rel_{sent}^{w}=\operatorname{rel}_{s}\left(S_{w}^{u}, I^{u}\right)$. We rank all the distractors by $Rel_{sent}^{w}$ and select the top 3 distractors as the final distractors.

\subsection{Quality Control}

To deliver a convincing evaluation method to existing VL models, we have humans verify the full validation set. We designed and deployed a user interface on Amazon Mechanical Turk platform and hired experienced turkers (with $\$12.6/hr$)to help verify the correctness of our questions and answers. Every image-question pair was cross-verified and corrected by five turkers

Having the image on the side, every turker would be first asked to verify the correctness of the question in terms of grammar or understanding. If the question is marked as incorrect or not understood, we would ask the turkers to help re-correct the question or skip it\footnote{For skipped data, the author would take a look at them to verify.}. Then the turkers would be provided with 7 answer choices (1 correct answer choice and 6 incorrect answer choices) and 2 additional choices of "None of the above" and "I do not know how to answer". 

Avoiding causing any prior biases in the turkers and resulting in false positives and false negatives, we do not inform turkers the number of correct answer choices and ask them to select all the ones they think are correct. If they cannot understand the visual scene or find a correct answer at all, they can even select "I do not know how to answer" or "None of the above". After selecting the answer choices, we also give the turkers options to go over every answer choice to re-correct it if there is any grammatical issue. In the end, if the turkers have selected "None of the above" before, they would be asked to created their own correct answer choices.

To ensure the correctness of the annotation interface, we first conduct many in-house experiments. After that, we also randomly select several turkers' annotations as pseudo groundtruths. We further evaluate other turkers' annotation against the pseudo groundtruths to ensure the agreement rate on selections. 

For an image-question pair, if turkers have different selections on the correct answer choices, we would avoid avoid using any answer choices selected as correct by any of the turker as a distractor. 

 When filtering the annotations, we ensure that every selected final distractor in ME cannot be selected by any of the turkers as correct before to avoid false negative. Further, when filtering every sample's annotations, among the five turkers, we ensure that the selected final correct answer choice should be selected by at least three of them to avoid false positive. If more than one answer choice is selected three times, we would compare and select the one that has the most selections.

\bibliography{anthology,custom}
\bibliographystyle{acl_natbib}

\clearpage

\end{document}